\documentclass[10pt,twocolumn,letterpaper]{article}

\usepackage[pagenumbers]{cvpr} %

\usepackage{graphicx}
\usepackage{amsmath}
\usepackage{amssymb}
\usepackage{booktabs}
\usepackage{epsfig}

\usepackage{booktabs,tabulary,multirow,overpic,xcolor}
\usepackage[caption=false]{subfig}
\usepackage[pagebackref=false, breaklinks=true, letterpaper=true, colorlinks,
            citecolor=citecolor, linkcolor=linkcolor, bookmarks=false]{hyperref}
\definecolor{citecolor}{HTML}{0071BC}
\definecolor{linkcolor}{HTML}{ED1C24}

\definecolor{boxinstcolor}{rgb}{0.58, 0.404, 0.741}
\definecolor{dextrcolor}{rgb}{1.00, 0.498, 0.055}

\usepackage[british,english,american]{babel}

\newcommand{\app}{\raise.17ex\hbox{$\scriptstyle\sim$}}

\def\x{\times}
\newcolumntype{x}[1]{>{\centering\arraybackslash}p{#1pt}}
\newlength\savewidth\newcommand\shline{\noalign{\global\savewidth\arrayrulewidth
  \global\arrayrulewidth 1pt}\hline\noalign{\global\arrayrulewidth\savewidth}}
\newcommand{\tablestyle}[2]{\setlength{\tabcolsep}{#1}\renewcommand{\arraystretch}{#2}\centering\footnotesize}
\makeatletter\renewcommand\paragraph{\@startsection{paragraph}{4}{\z@}
  {.5em \@plus1ex \@minus.2ex}{-.5em}{\normalfont\normalsize\bfseries}}\makeatother

\newcommand{\dt}[1]{\fontsize{7pt}{0.1em}\selectfont (#1)}

\usepackage[capitalize]{cleveref}
\crefname{section}{Sec.}{Secs.}
\Crefname{section}{Section}{Sections}
\Crefname{table}{Table}{Tables}
\crefname{table}{Tab.}{Tabs.}

\def\cvprPaperID{44} %
\def\confName{CVPR}
\def\confYear{2022}

\begin{document}

\title{Pointly-Supervised Instance Segmentation}

\author{
Bowen Cheng$^{1}$\thanks{Work done during an internship at Facebook AI Research.} \qquad Omkar Parkhi$^2$ \qquad Alexander Kirillov$^2$\vspace{.5em} \\[0mm]
$^1$UIUC  \qquad $^2$Facebook AI
}

\maketitle

\begin{abstract} 
   We propose an embarrassingly simple point annotation scheme to collect weak supervision for instance segmentation. In addition to bounding boxes, we collect binary labels for a set of points uniformly sampled inside each bounding box.
   We show that the existing instance segmentation models developed for full mask supervision can be seamlessly trained with point-based supervision collected via our scheme. Remarkably, Mask R-CNN trained on COCO, PASCAL VOC, Cityscapes, and LVIS with only 10 annotated random points per object achieves 94\%--98\% of its fully-supervised performance, setting a strong baseline for weakly-supervised instance segmentation. The new point annotation scheme is approximately 5 times faster than annotating full object masks, making high-quality instance segmentation more accessible in practice.
   
   Inspired by the point-based annotation form, we propose a modification to PointRend instance segmentation module. For each object, the new architecture, called Implicit PointRend, generates parameters for a function that makes the final point-level mask prediction. Implicit PointRend is more straightforward and uses a single point-level mask loss. Our experiments show that the new module is more suitable for the point-based supervision.
   \footnote{Project page: \scriptsize \url{https://bowenc0221.github.io/point-sup}}
\end{abstract}

\vspace{-4mm}
\section{Introduction}

The task of instance segmentation requires an algorithm to locate objects and delineate them with pixel-level binary masks. Manual annotation of object masks for training is significantly more complex and time-consuming than other forms of image annotation like image-level categories~\cite{zhou2018weakly,ahn2019weakly,zhu2019learning,cholakkal2019object,fan2018associating,liu2020leveraging,ge2019label} or per-object bounding boxes~\cite{khoreva2017simple,hsu2019weakly,arun2020weakly,tian2020boxinst}. For example, it takes on average 79.2 seconds per instance to create a polygon-based object mask in COCO~\cite{lin2014coco}, whereas a bounding box for an object can be annotated $\sim$11 times faster in only 7 seconds~\cite{papadopoulos2017extreme}.

\begin{figure}[!t]
  \vspace{-5mm}
  \centering
  \includegraphics[width=0.99\linewidth,trim={3.5cm 0cm 2cm 9.7cm},clip]{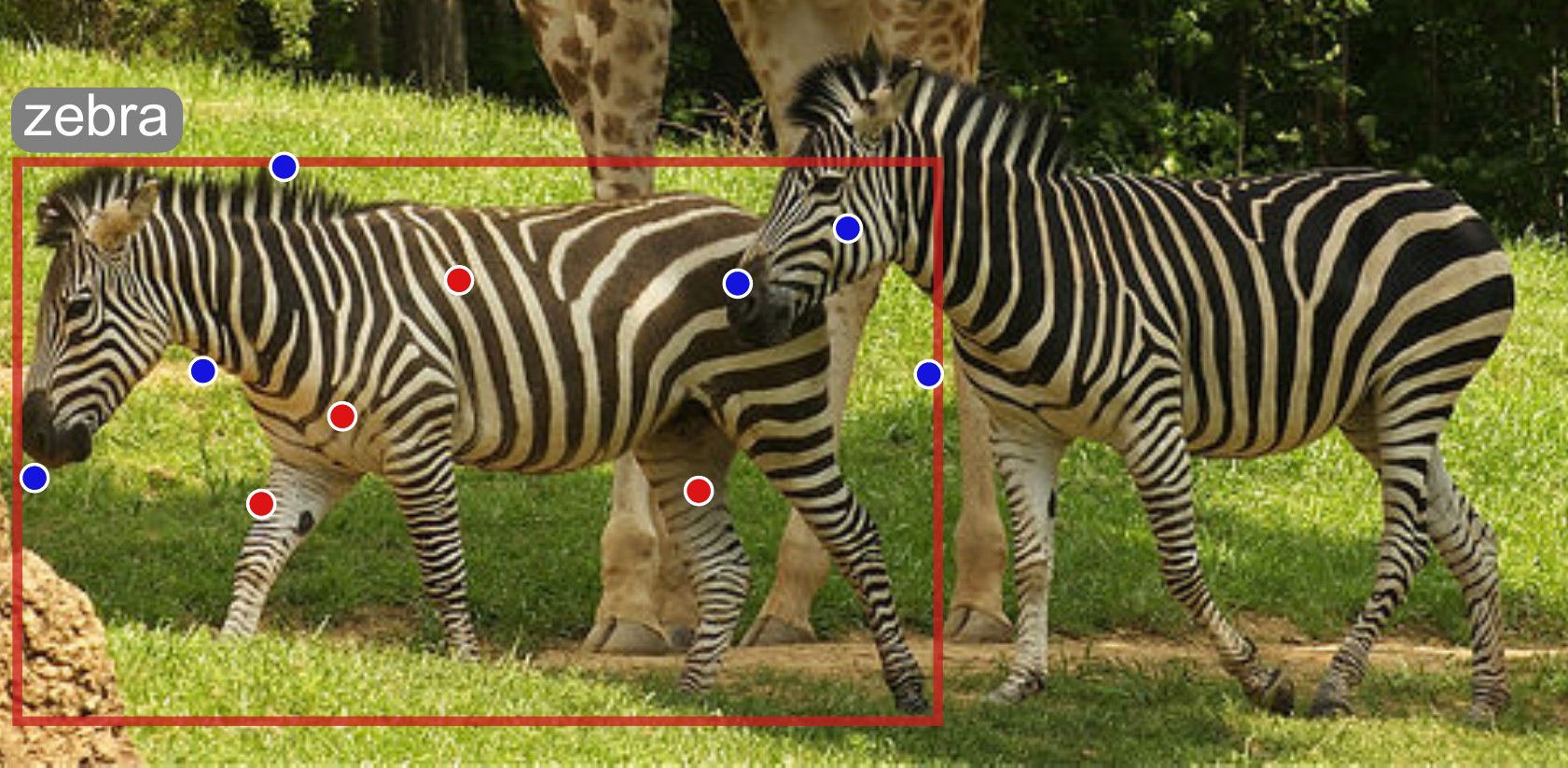}
  \caption{\textbf{Our point-based instance annotation scheme.} We collect object bounding boxes with points \emph{randomly sampled} inside each box and annotated as the object~(\textcolor{red}{red}) or background~(\textcolor{blue}{blue}). 
  Our experiments show that a bounding box and 10 annotated points per instance are approximately 5 times faster to collect than the standard object mask annotation and such ground truth is sufficient to train a standard model like Mask R-CNN~\cite{he2017mask} to achieve 94\%--98\% of its fully-supervised performance on various datasets.}
  \vspace{-4mm}
  \label{fig:teaser}
  \end{figure}

Weakly-supervised methods, that use easier to acquire ground truth annotation forms, make instance segmentation more accessible for new categories or scene types, as the efforts required to collect the data are lower. Such models showed a great progress on smaller datasets under a fixed annotation time budget~\cite{bearman2016s,laradji2020proposal}, however, they are still far behind fully-supervised methods for large-scale datasets like COCO. The recent BoxInst model~\cite{tian2020boxinst} outperforms previous weakly-supervised approaches with box supervision but achieves only 85\% of its fully-supervised counterpart performance on COCO. A natural question emerges: Is object mask training data necessary to get closer to the fully-supervised performance? And is there an easier to collect annotation form for the instance segmentation task?

Beyond bounding boxes and image-level categories, point clicks and squiggles are the other time-efficient annotation forms most commonly used in interactive segmentation scenarios~\cite{xu2016deep,liew2017regional,li2018interactive,maninis2018deep,benenson2019large}. Several semantic and instance segmentation methods directly use them for training~\cite{bearman2016s,lin2016scribblesup}. In our paper, we present a new instance segmentation annotation scheme that collects both bounding boxes and point-based annotation (see Figure~\ref{fig:teaser}). Unlike previous works where points are clicked by annotators, we follow a different process where points are sampled randomly inside an object bounding box and an annotator is asked to classify each point as the object or background. This point-based annotation scheme is simple and we empirically find such annotation to perform well on large-scale datasets.

With randomly sampled points, the annotation process can be easily simulated with existing instance segmentation ground truth. This property allows us to test the new annotation scheme and show its efficacy on multiple large-scale datasets without a major annotation effort that is usually a prerequisite for other annotation schemes~\cite{bearman2016s,lin2016scribblesup}.

The key advantage of the point-based annotation produced by our scheme is in its ability to seamlessly supervise instance segmentation models that directly predict object masks with no changes to their architectures or training pipelines. In our experiments, we train Mask R-CNN~\cite{he2017mask}, PointRend~\cite{kirillov2020pointrend}, and CondInst~\cite{tian2020conditional} with this point-based supervision. For each object, the standard mask loss is computed for ground truth points by interpolating mask predictions at these points. We show that the point-based annotation obtained via our scheme is applicable across different categories and scene types by simulating it on COCO~\cite{lin2016feature}, PASCAL VOC~\cite{everingham2015pascal}, LVIS~\cite{gupta2019lvis}, and Cityscapes~\cite{Cordts2016Cityscapes} datasets. Mask R-CNN trained with only 10 annotated points per object achieve 94\%--98\% of its fully-supervised performance on these datasets. In addition, we propose a simple point-based data augmentation strategy and explore self-training paradigm for point-based supervision to show that the gap to the full supervision can be further reduced. Finally, we show that point-based pre-training is matching mask-based in a transfer learning setup.

To analyze the performance/annotation time trade-off of the new annotation scheme, we created a simple labeling tool and measured that a trained human annotator classifies a point in 0.9 seconds on average.
This means, together with a bounding box annotation that can be done in 7 seconds via extreme point clicking~\cite{papadopoulos2017extreme}, the total annotation time for 10 points per object is 16 seconds ($7 + 10 \cdot 0.9$). This is 5 times faster than polygon-based mask annotation in COCO. For a new dataset, the point-based annotation from our scheme allows the standard models to achieve performance close to full supervision at a fraction of the full data collection time.

In our experiments, we observe that PointRend~\cite{kirillov2020pointrend} supervised with point-based annotations performs on par with Mask R-CNN~\cite{he2017mask} supervised in the same way, whereas with the standard full mask supervision it performs better. Inspired by this finding, we propose \textbf{Implicit PointRend}. Instead of a coarse mask prediction used in PointRend to provide region-level context to distinguish objects, for each object Implicit PointRend generates different parameters for a function that makes the final point-wise mask prediction. The new model is simpler than PointRend: (1) it does not require an importance point sampling during training and (2) it uses a single point-level mask loss instead of two mask losses. Implicit PointRend can be trained directly with point supervision without any intermediate prediction interpolation steps. Our experiments demonstrate that the new module outperforms PointRend with point supervision. %

\section{Related Works}

\paragraph{Fully-supervised instance segmentation} models that currently dominate popular COCO benchmark~\cite{lin2014coco} predict object masks directly. Mask R-CNN-based methods~\cite{he2017mask} make such predictions on a region level from features that correspond to detected bounding boxes, whereas methods like YOLACT++~\cite{yolact-plus-tpami2020} or CondInst~\cite{tian2020conditional} make image-level mask predictions. In both cases ground truth masks are directly used for supervision. Alternatively, bottom-up segmentation methods output masks indirectly, forming them from a set of predicted cues such as instance boundaries~\cite{kirillov2016instancecut}, energy levels~\cite{bai2016deep}, and offset vectors to center points~\cite{cheng2020panoptic}. We show that methods that use direct mask supervision can be trained with point-based annotation from our annotation scheme without significant modifications.

\paragraph{Weakly-supervised instance segmentation} methods usually use image-level category labels or bounding boxes. With image-level labels, such methods rely on segmentation proposals~\cite{uijlings2013selective,pont2016multiscale} either re-ranking them~\cite{zhou2018weakly} or generating pseudo-ground truth~\cite{ahn2019weakly, zhu2019learning, arun2020weakly}. These methods show promising results on smaller datasets, but no competitive results have been shown on a large-scale dataset like COCO~\cite{lin2014coco}. With bounding box supervision, SDI~\cite{khoreva2017simple} proposes a multi-stage training procedure that generates pseudo-ground truth with GrabCut~\cite{rother2004grabcut}. BBTP~\cite{hsu2019weakly} trains Mask R-CNN~\cite{he2017mask} using a bounding box tightness prior. Recently proposed BoxInst~\cite{tian2020boxinst} supervises the mask branch of CondInst~\cite{tian2020conditional} with a projection loss that forces horizontal/vertical lines inside bounding boxes to predict at least one foreground pixel and an affinity loss that forces pixels with similar colors to have the same label. BoxInst outperforms previous approaches, achieving an impressive 85\% of fully-supervised performance on COCO~\cite{lin2014coco}. 
In our work, we show that additional point-based supervision allows us to get much closer to the quality of fully-supervised models. 
In addition, our experiments suggest that our point-based annotation scheme has a better performance/annotation time trade-off for a wide variety of annotation efforts.

\paragraph{Point-based supervision} has been studied in a variety of tasks, including action localization~\cite{mettes2016spot,mettes2019pointly}, object detection~\cite{papadopoulos2017training,ren2020ufo}, object counting~\cite{laradji2018blobs}, and semantic segmentation~\cite{bearman2016s,qian2019weakly}. For instance segmentation, point clicks are often used in interactive pipelines~\cite{xu2016deep,liew2017regional,li2018interactive,maninis2018deep,benenson2019large}. While very powerful, such systems are more complex than our scheme, as they are trained with full supervision and require to repeatedly run an inference model during annotation. Laradji~\etal~\cite{laradji2020proposal} present a proposal-based instance segmentation method that uses a single point per instance as supervision. In contrast, we use multiple points and a bounding box annotation together. Furthermore, instead of collecting clicks, we randomly generate point locations and ask annotators to classify points as object or background.

\section{Pointly-Supervised Instance Segmentation}

\subsection{Annotation format and collection}
\label{sec:point:intro}

The format of point-based annotation collected via the new scheme is conceptually simple. In addition to the standard bounding box annotation, we assume $N$ points within each object bounding box to be labeled as the object \vs background (Figure~\ref{fig:teaser}).
We refer to this annotation as $\mathcal{P}_{N}$.

\paragraph{Random points instead of clicks.} Previous point-based annotation schemes collected annotators clicks~\cite{bearman2016s,qian2019weakly,laradji2020proposal,mettes2016spot,papadopoulos2017training,laradji2018blobs}. Such strategy, however, can lead to data with low variability as human clicks are often correlated~\cite{firestone2014please,clark2005coordinating}. Bearman~\etal~\cite{bearman2016s} confirm this by showing that training with randomly located ground truth points is superior for their semantic segmentation model than training with clicks. Following these findings we randomly sample point locations within each object bounding box and an annotator is asked to classify each point as the object or background.

\paragraph{Collection and simulation.} The annotation scheme for collecting $\mathcal{P}_{N}$ is straightforward. First, bounding boxes for objects are collected using any off-the-shelf solution~\cite{papadopoulos2017extreme,kuznetsova2020openimages,gygli2019efficient}. Next, for each object, $N$ random point locations within its bounding box are sequentially presented to an annotator for a binary object/background classification given the bounding box and object category label.

Unlike click-based schemes where human is involved in the point selection process, the new annotation scheme can be seamlessly simulated for datasets that have full instance segmentation ground truth by generating a bounding box for each instance mask and classifying randomly sampled points/pixels inside the box based on the corresponding positions in the ground truth mask. By simulating the annotation from our scheme, we are able to ablate its design and verify our results on a diverse set of large-scale datasets without time consuming data collection efforts.

\paragraph{Annotation time.} To measure the annotation time, we build a simple tool for our point-based annotation scheme 
(see the appendix for details). With this tool we annotated 100 objects from the COCO~\cite{lin2014coco} and LVISv1.0~\cite{gupta2019lvis} datasets, by an annotating contractor company with human annotators similar to Amazon Mechanical Turk (AMT), and found that it takes 0.9 seconds on average to classify a point. Note, that a bounding box can be annotated in 7 seconds using extreme points method~\cite{papadopoulos2017extreme}. Thus, the total annotation time of $\mathcal{P}_{N}$ per object is $7$ (bounding box) $ + $ $0.9 \cdot N$ (binary classification for $N$ points) seconds.

In our experiments we found that $\mathcal{P}_{10}$ provides a good trade-off between annotation time (16 seconds per object) and performance (94\%-98\% of fully-supervised performance). Such annotation is $\sim$5 times faster to label than a polygon-based instance mask in COCO that takes on average 79.2 seconds for a spotted object~\cite{lin2014coco}.

\paragraph{Annotation quality.} Collected point labels have 90\% agreement with COCO ground truth masks. In most cases, the misclassification occurs very close to object boundaries or in regions where the ground truth polygons are imprecise. The agreement is around 95\% when more accurate LVIS ground truth masks are used. Our experiments show no significant drop in performance if point labels are simulated with 95\% accuracy. In real world scenarios, such errors are usually fixed with a verification step~\cite{lin2014coco,gupta2019lvis,kuznetsova2020openimages}.

\begin{figure}[!t]
\centering
\bgroup
\def\arraystretch{0.2}
\setlength\tabcolsep{0.8pt}
\begin{tabular}{c}
\includegraphics[width=0.62\linewidth]{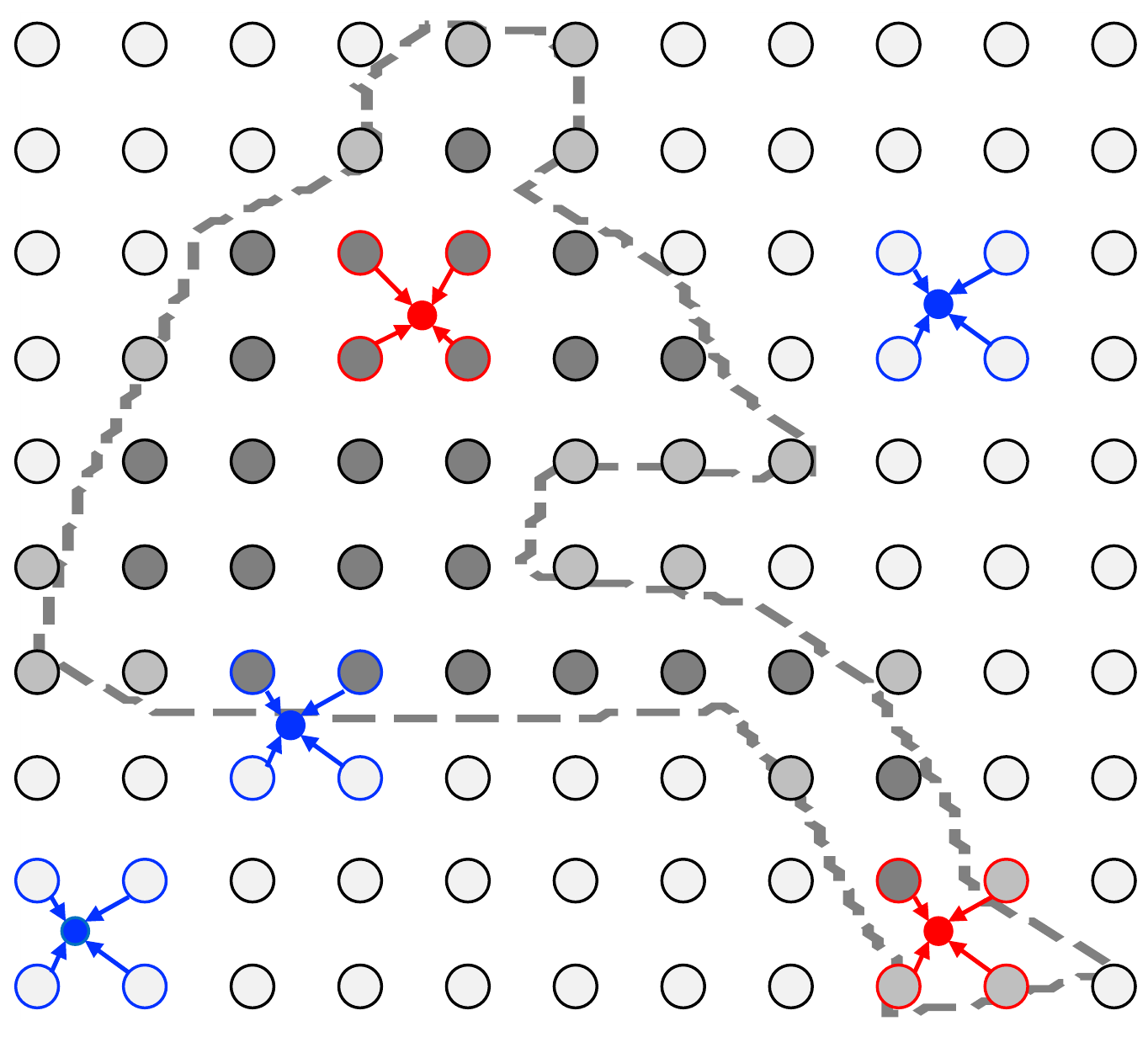} \\
\end{tabular} \egroup
\vspace{-2mm}
\caption{\textbf{Illustration of point supervision.} For a $10\x10$ prediction mask on the regular grid (darker color indicates foreground prediction), we get predictions at the exact locations of ground truth points (red and blue indicate foreground and background ground truth points respectively) with bilinear interpolation. Note, that the object contour line is only for the purpose of illustration.}
\vspace{-2mm}
\label{fig:mask_rcnn_with_point}
\end{figure}

\subsection{Training with points}
\label{sec:point:mask_rcnn}

The key advantage of such point-based annotation is its ability to seamlessly supervise off-the-shelf instance segmentation models that make mask predictions on a regular grid (\eg, a 28$\times$28 RoI prediction of Mask R-CNN~\cite{he2017mask} or an image-level per-pixel prediction of CondInst~\cite{tian2020conditional}). In a fully-supervised setting, these models are trained by extracting a matching regular grid of labels from ground truth masks. In contrast, with point supervision, we approximate predictions in the locations of ground truth points from the prediction on the grid using bilinear interpolation (see Figure~\ref{fig:mask_rcnn_with_point}). Once we have predictions and ground truth labels at the same points, a loss can be applied in the same way as with full supervision and its gradients will be propagated through bilinear interpolation. Note, that such supervision does not require any changes to the architecture. In our experiments we use cross-entropy loss on points, however other losses like dice loss~\cite{milletari2016v} can be used as well.

For region-based models, some ground truth points may lay outside of predicted boxes and we choose to ignore such points during training. Contrarily, for models that yield image-level masks, additional background points can be sampled from the outside of ground truth boxes. However, in this paper we study the most basic setup where only $N$ annotated ground truth points are used to train all models.

\paragraph{Data augmentation} during training is a crucial component of modern segmentation models. The point-based annotation is compatible with all common augmentations like scale jitter, crop, or horizontal flip of an input image. In our experiments, we observe that the gap in performance between point and full mask supervision is larger for longer training (3$\x$ schedule~\cite{wu2019detectron2}) and models with higher-capacity backbones (\eg, ResNeXt-101~\cite{xie2017aggregated}). We hypothesize that this is caused by a reduced variability of training data when only a few points are available and propose a extremely simple \emph{point-based} data augmentation strategy: instead of using all available ground truth points for a box at each iteration, we subsample half of all available points at every training iteration (5 points for $\mathcal{P}_{10}$ ground truth). In the next section we show that our augmentation strategy improves performance for higher-capacity models.

\subsection{Experiments with point supervision}
\label{sec:point:experiments}

In this section we first ablate the design of the new annotation scheme on the COCO dataset~\cite{lin2014coco} using Mask R-CNN~\cite{he2017mask}. Then, we demonstrate the effectiveness of the point-based supervision across 4 different datasets and show that it is applicable to a diverse set of instance segmentation models. Finally, we explore the annotation time and performance trade-off of the new annotation scheme. 

\paragraph{Implementation details.} 

We use Mask R-CNN~\cite{he2017mask} with a ResNet-50-FPN~\cite{he2016deep,lin2016feature} backbone and the default training schedules and parameters from Detectron2~\cite{wu2019detectron2} for each dataset ($3\x$ schedule is used for COCO~\cite{lin2014coco}). Apart from the mask branch loss, there are no other differences between full mask and the point-based supervision in our experiments. Unless stated otherwise, we do not use point-based data augmentations described in section~\ref{sec:point:mask_rcnn}.

\paragraph{Datasets.} The main dataset in our experiments is COCO~\cite{lin2014coco} which has 118k training and 5k validation images with 80 common categories. We also conduct experiments using: PASCAL VOC dataset~\cite{everingham2015pascal} which is smaller than COCO with $\sim$10k images and 20 categories; Cityscapes~\cite{Cordts2016Cityscapes} ego-centric street-scene dataset that has 2975 train and 500 validation high-resolution images with high-quality instance-level annotations for 9 categories; LVISv1.0~\cite{gupta2019lvis} that uses the same set of images as COCO but has more than 1000 categories annotated in a federated fashion. We use the standard evaluation metrics: AP50~\cite{everingham2015pascal} for PASCAL VOC and AP~\cite{lin2014coco} for all other datasets. 

For all datasets in our experiments, we simulate point-based ground truth by randomly sampling pixels inside each ground truth bounding box and selecting their labels based on the corresponding ground truth mask.  %

\subsubsection{Ablation of the annotation design}

\begin{figure}[!t]
  \centering
  \includegraphics[width=0.8\linewidth]{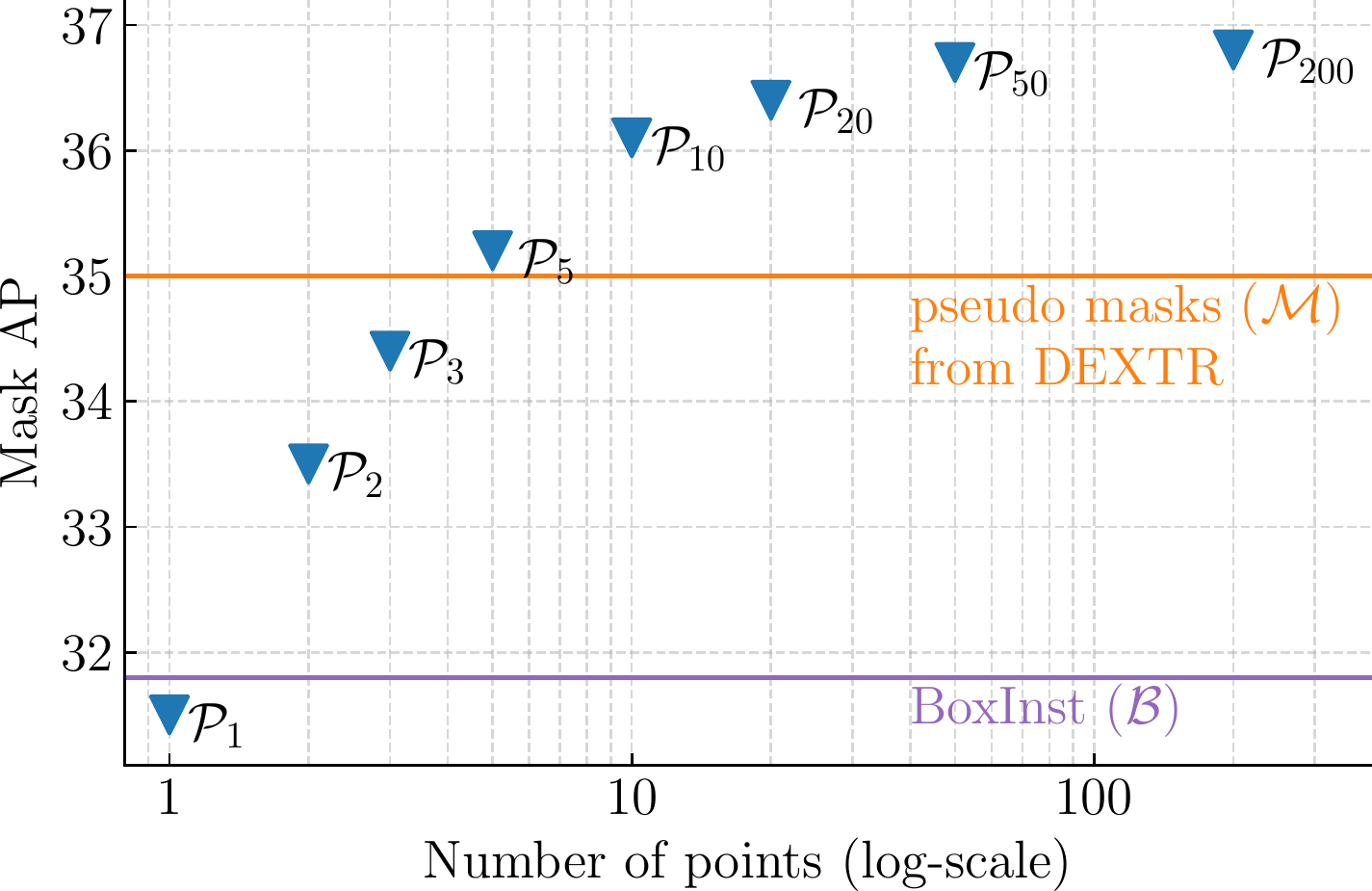}
  \vspace{-2mm}
  \caption{\textbf{Training with a different number of points.} Mask R-CNN~\cite{he2017mask} with a ResNet-50-FPN backbone trained on COCO with as few as 10 labeled points per instance ($\mathcal{P}_{10}$) achieves 36.1 mask AP with diminishing returns from more labeled points.
  We provide two baselines: (1) \textcolor{boxinstcolor}{BoxInst}~\cite{tian2020boxinst}, the best current model that uses box supervision ($\mathcal{B}$) only, and (2) Mask R-CNN trained with pseudo ground truth generated by an interactive segmentation method \textcolor{dextrcolor}{DEXTR}~\cite{maninis2018deep} that uses extreme clicks as its input.
  }\vspace{-3mm}
  \label{fig:coco_num_points_maskrcnn}
  \end{figure}

\paragraph{Number of points.} In Figure~\ref{fig:coco_num_points_maskrcnn} we demonstrate the performance of a Mask R-CNN trained with a varying number of labeled points per instance. The performance rapidly improves with the number of annotated points going up to tens with diminishing returns thereafter. While 20 points ($\mathcal{P}_{20}$) is approximately 0.3 AP better than 10 point supervision ($\mathcal{P}_{10}$), it takes twice as long to annotate. In what follows, we use the 10 points supervision which is $\sim$5 times faster than polygon-based mask annotation. Two baselines in Figure~\ref{fig:coco_num_points_maskrcnn} are: (1) \textcolor{boxinstcolor}{BoxInst}~\cite{tian2020boxinst}, the best current model that uses box supervision ($\mathcal{B}$) only, and (2) Mask R-CNN trained with masks generated by \textcolor{dextrcolor}{DEXTR}~\cite{maninis2018deep}, an interactive segmentation method that predicts a mask from 4 extreme clicks. For this baseline, DEXTR is trained in PASCAL VOC~\cite{everingham2015pascal} and we simulated extreme clicks on COCO as its input.

\begin{figure*}[!t]
\centering
\includegraphics[width=0.93\linewidth,trim={0.0cm 0.0cm 0.0cm 1.2cm},clip]{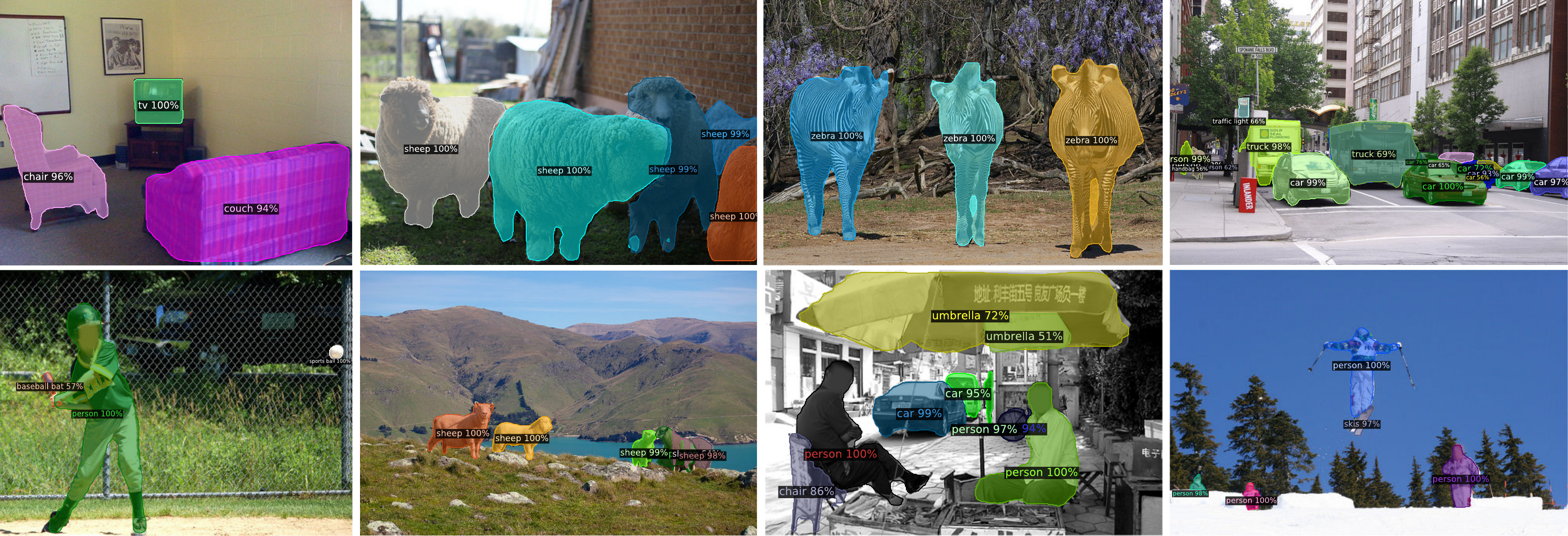}
\vspace{-3mm}
\caption{\textbf{Mask R-CNN trained with 10 points per object ($\mathcal{P}_{10}$)} on COCO. The model uses a ResNet-50-FPN~\cite{he2017mask,lin2016feature} backbone and is trained with $3\x$ schedule~\cite{wu2019detectron2} ($36.1$ AP). We show predictions with confidence scores greater than 0.5.}
\label{fig:mask_rcnn_R50_visualization}
\end{figure*}

\begin{table*}[t]
  \vspace{-5mm}
  \centering
  \subfloat[COCO \texttt{val2017}~\cite{lin2014coco}.\label{tab:dataset:coco}]{
  \tablestyle{5pt}{1.1}\begin{tabular}{c|x{34}}
   supervision & AP \\
  \shline
   $\mathcal{M}$ & 37.2 \phantom{\dt{99\%}} \\
   $\mathcal{P}_{10}$ & 36.1 \dt{97\%} \\
  \end{tabular}}\hspace{3mm}
  \subfloat[PASCAL VOC \texttt{val}~\cite{everingham2015pascal}.\label{tab:dataset:voc}]{
  \tablestyle{5pt}{1.1}\begin{tabular}{c|x{34}}
   supervision & AP$_{50}$ \\
  \shline
   $\mathcal{M}$ & 66.3 \phantom{\dt{99\%}} \\
   $\mathcal{P}_{10}$ & 64.2 \dt{97\%} \\
  \end{tabular}}\hspace{3mm}
  \subfloat[Cityscapes \texttt{val}~\cite{Cordts2016Cityscapes}.\label{tab:dataset:cityscapes}]{
  \tablestyle{5pt}{1.1}\begin{tabular}{c|x{34}}
   supervision & AP \\
  \shline
   $\mathcal{M}$ & 32.7 \phantom{\dt{99\%}} \\
   $\mathcal{P}_{10}$ & 30.7 \dt{94\%} \\
  \end{tabular}}\hspace{3mm}
  \subfloat[LVISv1.0 \texttt{val}~\cite{gupta2019lvis}.\label{tab:dataset:lvis}]{
  \tablestyle{5pt}{1.1}\begin{tabular}{c|x{34}}
   supervision & AP \\
  \shline
   $\mathcal{M}$ & 22.8 \phantom{\dt{99\%}} \\
   $\mathcal{P}_{10}$ & 21.5 \dt{94\%} \\
  \end{tabular}}\vspace{-3mm}
\caption{\textbf{Mask R-CNN with full mask ($\mathcal{M}$) \vs new point supervision ($\mathcal{P}_{10}$).} For each object only 10 labeled points are used. For all four datasets with different properties, pointly-supervised models are within 2 AP (94\% -- 97\%) from their fully-supervised counterparts.}
\label{tab:dataset} %
\end{table*}

\begin{table}[t]
  \vspace{-2mm}
  \centering
  \tablestyle{4pt}{1.1}\begin{tabular}{c c | x{39}x{39}x{39}}
   supervision & point aug. & R50 AP & R101 AP & X101 AP \\
  \shline
  $\mathcal{M}$ & - & 37.2 \phantom{\dt{96\%}} & 38.6 \phantom{\dt{96\%}} & 39.5 \phantom{\dt{96\%}} \\
  \multirow{2}{*}{$\mathcal{P}_{10}$}  & & 36.1 \dt{97\%} & 37.8 \dt{98\%} & 38.1 \dt{96\%} \\
    & \checkmark & 36.0 \dt{97\%} & 37.8 \dt{98\%} & 38.5 \dt{97\%} \\
  \end{tabular}\vspace{-3mm}
\caption{\textbf{Mask R-CNN with higher-capacity backbones and point-based augmentation} on COCO. We use ResNet-50 (R50), ResNet-101 (R101), and ResNeXt101-32$\times$8 (X101) backbones~\cite{he2016deep,xie2017aggregated} with FPN~\cite{lin2016feature}. The simple point-based data augmentation (\emph{point aug.})  proposed in section~\ref{sec:point:mask_rcnn} improves performance (+0.4 AP) for the higher-capacity X101 model.
}
\label{tab:overfitting}
\end{table}

\paragraph{Sensitivity to point locations.} We generate 5 different simulated versions of $\mathcal{P}_{10}$ ground truth annotation for the COCO \texttt{train2017} set using different random seeds for the uniform point sampling procedure. Training the same Mask R-CNN~\cite{he2017mask} model with these dataset versions, we observe very low 0.1 AP variation on the COCO \texttt{val2017} set. This result suggests that the point-based annotation from our annotation scheme is very robust with respect to the locations of sampled points. In what follows we conduct experiments with a single set of sampled points per dataset.

\paragraph{Sensitivity to the annotation quality.} We study the sensitivity to the quality of point-based annotation by changing correct object/background labels for random 5\% of points in a simulated ground truth $\mathcal{P}_{10}$ point annotation for COCO. The performance of a Mask R-CNN model trained on the tampered data is only 0.2 AP worse than the model trained on the clean data (35.9 AP \vs 36.1 AP). In a more challenging setup, instead of random points, we set wrong labels to 5\% of points that are closest to object boundaries. In this case, the drop in performance is slightly larger -- 0.4 AP (35.7 AP \vs 36.1 AP). Our experiment shows that such point-based annotation does not require perfectly accurate point labeling procedure which reduces the need for additional verification steps in a real-world annotation pipeline.

\subsubsection{Main results}

In Table~\ref{tab:dataset} we compare 10 points supervision ($\mathcal{P}_{10}$) to full mask supervision ($\mathcal{M}$) for a Mask R-CNN model with a ResNet-50-FPN~\cite{he2016deep,lin2016feature} backbone on four different datasets. Without any modifications to either the architecture or the default training algorithm/hyper-parameters set in Detectron2~\cite{wu2019detectron2}, Mask R-CNN trained with $\mathcal{P}_{10}$ achieves 94\% -- 97\% of the fully-supervised ($\mathcal{M}$) model performance. This result shows that such point supervision is widely applicable to datasets at different scales (\eg, from 20 categories in PASCAL VOC~\cite{everingham2015pascal} to more than 1,000 categories in LVISv1.0~\cite{gupta2019lvis}) and in different domains (common objects of COCO~\cite{lin2014coco} or street scenes of Cityscapes~\cite{Cordts2016Cityscapes}). In Figure~\ref{fig:mask_rcnn_R50_visualization}, we visualize predictions of a Mask R-CNN model supervised with points on COCO.

\begin{table}[t]
  \vspace{-6mm}
  \centering
  \subfloat[Mask R-CNN~\cite{he2017mask}.\label{tab:model:mask_rcnn}]{
  \tablestyle{2pt}{1.1}\begin{tabular}{c|x{34}}
   supervis. & AP \\
  \shline
   $\mathcal{M}$ & 37.2 \phantom{\dt{96\%}} \\
   $\mathcal{P}_{10}$ & 36.1 \dt{97\%} \\
  \end{tabular}}\hspace{0.9mm}
  \subfloat[CondInst~\cite{tian2020conditional}.\label{tab:model:condinst}]{
  \tablestyle{1pt}{1.1}\begin{tabular}{c|x{34}}
   supervis. & AP \\
  \shline
   $\mathcal{M}$ & 37.5 \phantom{\dt{96\%}} \\
   $\mathcal{P}_{10}^*$ & 35.7 \dt{95\%} \\
  \end{tabular}}\hspace{0.9mm}
  \subfloat[PointRend~\cite{kirillov2020pointrend}.\label{tab:model:pointrend}]{
  \tablestyle{1pt}{1.1}\begin{tabular}{c|x{34}}
    supervis. & AP \\
  \shline
   $\mathcal{M}$ & 38.3 \phantom{\dt{96\%}} \\
   $\mathcal{P}_{10}^*$ & 35.7 \dt{93\%} \\
  \end{tabular}}\vspace{-3mm}
\caption{\textbf{Different models trained with point supervision} on COCO. All models use a ResNet-50-FPN backbone~\cite{he2016deep,lin2016feature}. ``$*$'':~We train CondInst and PointRend models with point-based augmentation which improves their performance. While all three models achieve similar performance with point supervision, we observe that the gap between mask and point supervision is the largest for PointRend. We explore PointRend performance and propose an improved version of the module in the next section.}
\label{tab:model}
\end{table}

\paragraph{Higher-capacity backbones and point-based data augmentation.} In Table~\ref{tab:overfitting} we compare full and 10 points supervision for Mask R-CNN with different backbones on COCO. The simple point-based augmentation strategy described in section~\ref{sec:point:mask_rcnn} does not significantly change the performance of the smaller model. However, for higher-capacity ResNeXt-based model, the point-based augmentation effectively improves performance by 0.4 AP.

\paragraph{Different models.} Point-based supervision can be applied to a diverse set of models that use a per-pixel mask loss. In our experiments we use two methods in addition to Mask R-CNN: CondInst~\cite{tian2020conditional} that makes image-level predictions without an RoI pooling operation\footnote{
  As described in section~\ref{sec:point:mask_rcnn}, for CondInst~\cite{tian2020conditional} we use per-point cross entropy for given ground truth points \emph{without interpolation} and do not supervise predictions outside of the ground truth boxes. Therefore, we need to filter out any mask prediction outside of predicted boxes. A more sophisticated point selection strategy can be used to avoid it, however, this is not the focus of our paper.
  }
  and PointRend~\cite{kirillov2020pointrend} that makes point-level prediction to iteratively achieve a high resolution. We use the same ResNet-50-FPN~\cite{he2016deep,lin2016feature} backbone for all models and report results on COCO. Unlike Mask R-CNN, both CondInst and PointRend benefit from point-based augmentation even with a smaller backbone; thus, we report their results using the augmentation. In Table~\ref{tab:model} we observe that all three models achieve similar performance with point supervision. However, the gap between mask and point supervision for PointRend is significantly larger. The coarse mask head in PointRend makes prediction with low 7$\times$7 resolution making direct point-based training inaccurate. Inspired by this observation, in section~\ref{sec:ipr} we present a new streamlined version of PointRend module with a single mask loss which can be trained with either mask or point supervision.

\paragraph{Self-training with points.} The gap between point and full mask supervision can be further decreased using self-training paradigm~\cite{scudder1965probability,yarowsky1995unsupervised,riloff2003learning,xie2020self,chen2020naive,zoph2020rethinking}. After training a ResNet-50-FPN based Mask R-CNN with 10 points supervision on COCO (36.1 AP), we collect pseudo-ground truth masks by running inference on COCO train data with ground truth boxes. Mask R-CNN trained on these pseudo-ground truth masks without any changes in the training recipe, achieves 36.7 AP (+0.6 AP) or 98\% of fully-supervised Mask R-CNN trained with 6$\times$ schedule that matches the training time of our self-training setup.

\begin{table}[t]
  \vspace{-6mm}
  \centering
  \subfloat[Mask pre-training.\label{tab:transfer:mask_pretrain}]{
  \tablestyle{3pt}{1.1}\begin{tabular}{c|x{38}x{38}}
   supervis. & ImageNet & COCO-$\mathcal{M}$ \\
  \shline
   $\mathcal{M}$ & 66.3 \phantom{\dt{96\%}} & 74.5 \phantom{\dt{96\%}} \\
   $\mathcal{P}_{10}$ & 64.6 \dt{97\%} & 74.0 \dt{99\%} \\
  \end{tabular}}\hspace{3mm}
  \subfloat[Point pre-training.\label{tab:transfer:point_pretrain}]{
  \tablestyle{3pt}{1.1}\begin{tabular}{l|x{38}}
    pre-training & AP$_{50}$ ($\mathcal{M}$) \\
  \shline
   COCO-$\mathcal{M}$ & 74.5 \phantom{\dt{100\%}} \\
   COCO-$\mathcal{P}_{10}$ & 74.5 \dt{100\%} \\
  \end{tabular}}\vspace{-3mm}
\caption{\textbf{Pre-training with different supervision} and fine-tuning on PASCAL VOC. All models use a ResNet-50-FPN backbone~\cite{he2016deep,lin2016feature}. (a) Pre-training on datasets with mask annotation (\eg, COCO-$\mathcal{M}$) closes the gap between point-based supervision and full supervision on downstream dataset. (b) Point pre-training (COCO-$\mathcal{P}_{10}$) is as effective as mask pre-training (COCO-$\mathcal{M}$).}
\label{tab:transfer}
\end{table}

\paragraph{Transfer learning with points.} Pre-training on a large-scale dataset (\eg, COCO) followed by fine-tuning is a widely used paradigm to improve performance on a smaller dataset (\eg, PASCAL VOC). We study point-based annotation under two realistic scenarios: (1) if we have access to mask annotation in the large-scale dataset, Table~\ref{tab:transfer:mask_pretrain} shows mask pre-training closes the gap between point-based supervision and full supervision on downstream dataset; (2) our point-based annotation scheme is enough to collect a larger dataset for pre-training purpose, as Table~\ref{tab:transfer:point_pretrain} suggests that point pre-training is as effective as mask pre-training.

\subsubsection{Annotation time and performance trade-off.}

We compare the new point-based annotation scheme with other annotation collection processes for instance segmentation under \emph{the same annotation budget} which we measure as the time required to label training data. For this comparison we use identical Mask R-CNN~\cite{he2017mask} models for full mask and point supervision. Whereas, for bounding box supervision we use BoxInst~\cite{tian2020boxinst}, a recently proposed instance segmentation model that shows the best performance on COCO among existing weakly-supervised methods with bounding box supervision. Note, that BoxInst is based on CondInst~\cite{tian2020conditional} which performs on par or better than Mask R-CNN with full mask and point supervision. For all models we use the standard ResNet-50-FPN backbone~\cite{he2016deep,lin2016feature}.

\begin{figure}[!t]
  \centering
  \includegraphics[width=0.8\linewidth]{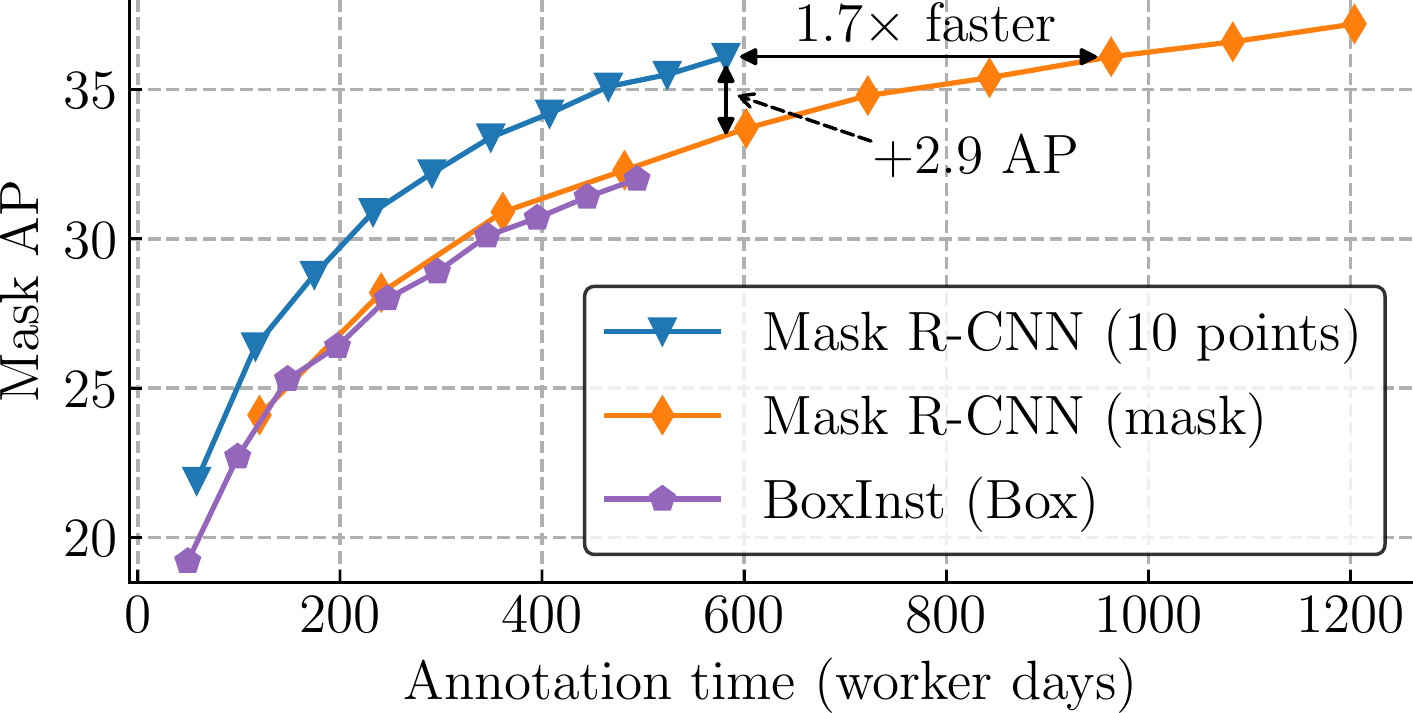}
  \vspace{-2mm}
\caption{\textbf{AP \vs annotation time for different supervision types} on COCO \texttt{val2017}. To match annotation times between different supervision forms, we train a Mask R-CNN model using from 10\% to 100\% of COCO \texttt{train2017}. Observe that Mask R-CNN trained with the new point-based scheme significantly outperforms models trained with both full mask supervision and weak bounding box supervision under the same computation budget.}
\label{fig:coco_ap_vs_annotation_time}
\end{figure}

\paragraph{COCO annotation timings.} An annotation collection for instance segmentation is a multi-stage process. For example, in the COCO annotation pipeline~\cite{lin2014coco}, annotators first spot and categorize objects by pointing at them, and then the spotted objects are annotated with polygon-based masks.
As discussed in section~\ref{sec:point:intro}, for a spotted object in COCO, it takes on average 7 seconds to annotate its bounding box~\cite{papadopoulos2017extreme}, 16 seconds to annotate the box and 10 points inside, and 79.2 seconds for polygon-based mask annotation~\cite{lin2014coco}. For each annotation form we approximate the total time to label COCO as the time required for the categorization and spotting stages (reported by COCO creators~\cite{lin2016feature}, see the appendix) plus the time required to annotate all spotted objects with the corresponding annotation form. In Figure~\ref{fig:coco_ap_vs_annotation_time} we match the annotation times between different supervision forms by training a Mask R-CNN model using from 10\% to 100\% of COCO \texttt{train2017} data. 
Our analysis shows that under the same annotation budget, Mask R-CNN trained with points significantly outperforms models trained with the other supervision forms.%

\section{Implicit PointRend Model}\label{sec:ipr}

Intuitively, point-based annotation from the new scheme should suite well to the PointRend module~\cite{kirillov2020pointrend} that yields a mask for a detected object using point-wise predictions. However, as we observed in Table~\ref{tab:model}, the gap between full mask and point supervision is larger for PointRend than for the other studied methods. While PointRend-based model trained with full mask supervision outperforms Mask R-CNN~\cite{he2017mask}, it performs worse than the baseline with point supervision. We hypothesize that the gap in performance is due to the coarse mask head of the standard PointRend module that outputs a $7 \times 7$ mask prediction and has its own mask loss. With such low resolution if two ground truth points are close to each other but have different labels, then this head does not get a reliable training signal.

Inspired by this observation, we propose a simplified version of the PointRend module which we name Implicit PointRend. For each detected object, instead of a coarse mask prediction, the new architecture predicts parameters for a point head function that can make a point-wise object mask prediction for any point given its position and corresponding image features. The new module has a single point-level mask loss which simplifies its implementation in comparison with PointRend. In what follows we describe the new design in detail and compare it with PointRend using both full mask and point supervision. 

\subsection{Point-wise representation and point head}

\paragraph{PointRend~\cite{kirillov2020pointrend}} constructs feature representation for a point concatenating two feature types: (1) a fine-grained point representation extracted from an image-level feature map (e.g. an FPN~\cite{lin2016feature} level feature map) in the exact position of the point via bilinear interpolation and (2) a coarse mask prediction for the point which provides region-specific information. The coarse mask prediction is made by a separate head that takes in RoI features and return a low resolution ($7 \times 7$) mask prediction. Given this point representation, a small MLP network, called point head, makes mask prediction for the exact point location. Note, that in the PointRend model, the parameters of the point head are shared across all points and detected instances.

\begin{figure}[!t]
\centering
\bgroup
\def\arraystretch{0.2}
\setlength\tabcolsep{0.8pt}
\begin{tabular}{c}
\includegraphics[width=0.93\linewidth,trim={4cm 11.5cm 12cm 0.5cm},clip]{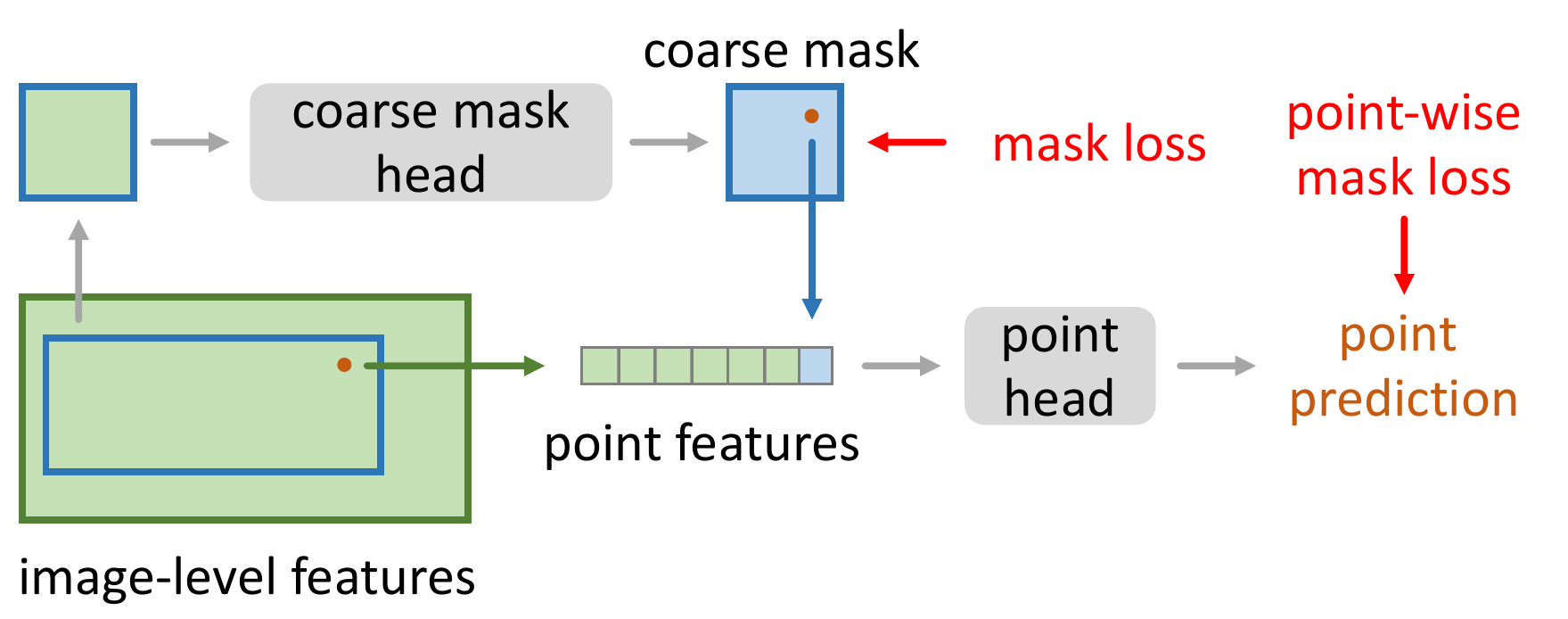} \\
\small (a) PointRend~\cite{kirillov2020pointrend}\\[3mm]
\includegraphics[width=0.93\linewidth,trim={4cm 11.5cm 12cm 0.5cm},clip]{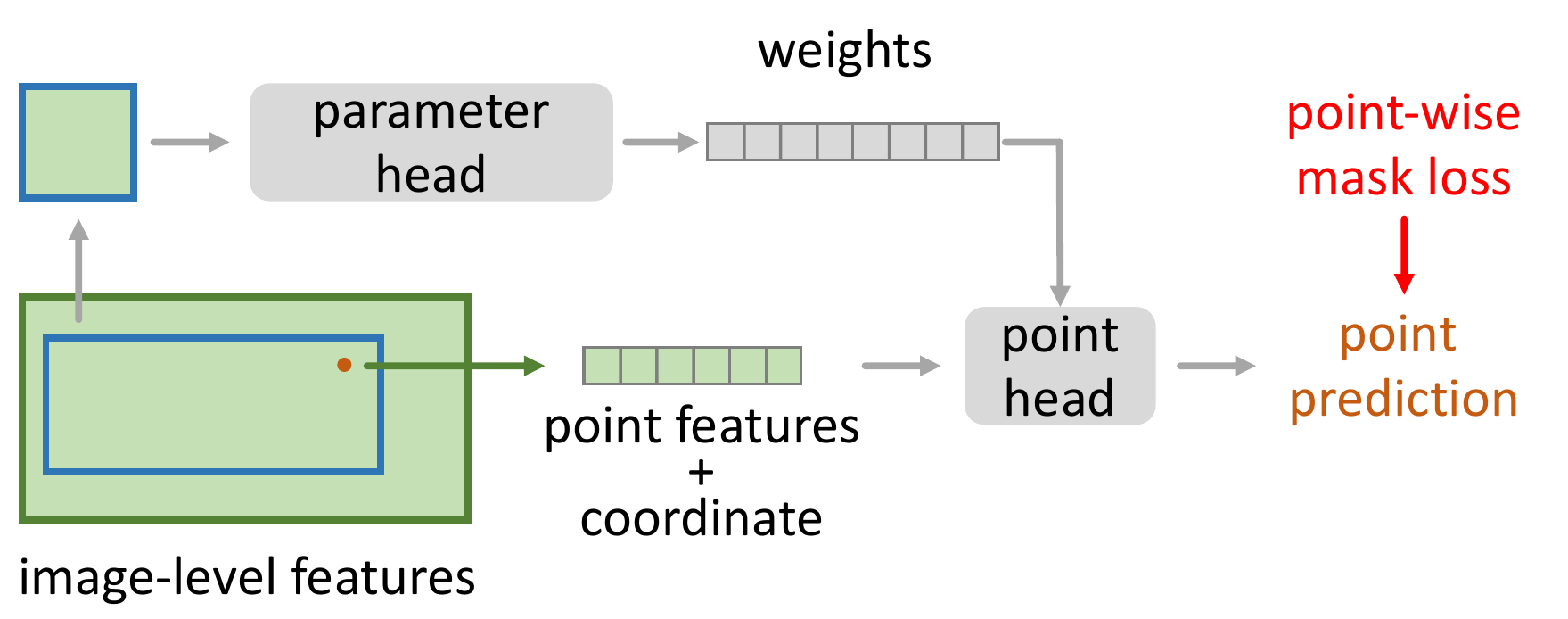} \\
\small (b) Implicit PointRend\\
\end{tabular} \egroup
\vspace{-2mm}
\caption{\textbf{PointRend~\cite{kirillov2020pointrend} \vs Implicit PointRend architectures.} Instead of a coarse mask prediction used in PointRend to provide region-level context to distinguish objects, Implicit~PointRend generates different parameters of the point head for each detected object. The point head makes prediction at any location independently taking in fine-grained point features and the information about the coordinate of this location relative to the bounding box. The same subdivision mask rendering algorithm~\cite{kirillov2020pointrend} is used in both to output high resolution mask with point-wise predictions.}
\label{fig:arch}
\end{figure}

\paragraph{Implicit PointRend} uses the point head to make point-wise mask predictions as well. However, instead of relying on coarse mask predictions to distinguish different instances, the new model generates different parameters of the point head for each instance similarly to the dynamic head approach of CondInst~\cite{tian2020conditional}. The parameters of the point head implicitly represent the mask of an objects and resembles implicit functions used by 3D community~\cite{jiang2020local,mescheder2019occupancy,genova2020local}. To make a mask prediction the point head takes in the coordinate of the point relative to the bounding box it belongs to. In addition, we use the same fine-grained point features as in the original PointRend design. Note, that the new Implicit PointRend module does not require a coarse mask prediction and, therefore, can be trained with a single mask loss applied to the output of the point head. We compare overall design of the new module with PointRend in Figure~\ref{fig:arch}.

\subsection{Point selection for inference and training}

\paragraph{Inference.} Implicit PointRend follows the same point selection strategy as PointRend during inference, \ie adaptive subdivision~\cite{whitted2005improved}. We start from predictions generated by our point head using a $28 \times 28$ uniform grid of points. Then, we gradually upsample it by a factor of 2 to $224 \times 224$ resolution. In each subdivision step, we select $N$ most uncertain points after interpolation and replace them with the predictions from the point head at these locations. Following the original PointRend implementation, we set $N$ to $28^2$.

\paragraph{Mask supervision ($\mathcal{M}$) training.} During training, PointRend selects more points from the uncertain regions of the coarse mask prediction via an importance sampling strategy. In our experiments we find that the Implicit PointRend model performs on par with PointRend while using a much simpler uniform point sampling strategy.

\paragraph{Point supervision ($\mathcal{P}_{10}$) training.} Implicit PointRend is naturally suited for point supervision, where we simply select points with ground truth annotation. Similarly to Mask R-CNN, we ignore points outside predicted boxes.

\subsection{Implementation details}
\label{sec:implicit_pointrend:intro:implementation_details}

We set all hyper-parameters following the original PointRend setup~\cite{kirillov2020pointrend}. The point head is an MLP with 3 hidden layers of 256 channels, ReLU activations~\cite{nair2010rectified} in hidden layers and the sigmoid activation applied to its output. To make a fair comparison, the head that generates parameters for the point head has the same architecture as the coarse mask head of PointRend~\cite{kirillov2020pointrend}. We observe however that Implicit PointRend performs as well with a smaller parameter head such as the standard box head architecture~\cite{Ren2015a}. We expect that a better design can bring further improvement.

As the input for the point head we use fine-grained features extracted from \texttt{p2} level of FPN~\cite{lin2016feature} that has 256 channel dimension. We use the random Fourier positional encoding~\cite{tancik2020fourier} to represent the point $(x, y)$ location relative to the center of the bounding box. We found that the model that uses positional encoding performs better than the one that uses $(x,y)$ coordinates directly. Please see the appendix for more detail and ablation experiments.

Implicit PointRend uses a single per-point mask loss (binary cross entropy) applied to the outputs of the point head. In addition, following the common setup for implicit function models~\cite{tancik2020fourier}, we add $l_2$ loss on predicted parameters of the point head to avoid predicted parameters become unbounded. This loss plays the role of weight decay which is otherwise absent for the dynamic parameters. In our experiments, we set the $l_2$ loss weight to 1e-5.

\subsection{Experimental evaluation}
\label{sec:implicit_pointrend:experiments}

To evaluate Implicit PointRend, we use COCO~\cite{lin2014coco} and 3$\x$ schedule~\cite{wu2019detectron2} to avoid underfitting. All hyper-parameters for the Implicit PointRend module are the same as in PointRend~\cite{kirillov2020pointrend}. In all experiments we attach the modules as a mask head to Faster R-CNN~\cite{Ren2015a}. Unless specified, we use a ResNet-50-FPN~\cite{he2016deep,lin2016feature} backbone.

In our experiments, we observe that Implicit PointRend performance significantly increases even for smaller models if the point sub-sampling augmentation described in section~\ref{sec:point:mask_rcnn} is used (see the appendix for an ablation). We did not observe the same behavior with the standard PointRend module and hypothesize, that the implicit representation of an object mask is more prone to overfitting than the coarse mask-based representation due to its higher capacity. To make a fair comparison, in this section we use the point-based augmentation for all models supervised with points.

\paragraph{Main results.} We compare the new Implicit PointRend and PointRend in Table~\ref{tab:coco_arch} with both full mask and point supervision. Unlike PointRend, our new module does not use any importance point sampling strategy and has only one mask loss. The more straightforward Implicit PointRend module performs on par with PointRend trained with full mask supervision (AP ($\mathcal{M}$)). Moreover, the new module significantly outperforms the baseline in case of point supervision (AP ($\mathcal{P}_{10}$)). This observation underscores unique challenges of the point supervision and suggests that new model designs may be needed to fully uncover the potential of the point-based annotation form.  %

Similar to the other methods studied in section~\ref{sec:point:experiments}, Implicit PointRend-based model supervised with points achieves 96\% of its full mask supervision performance. In Table~\ref{tab:ipr:ablation} we compare Implicit PointRend trained with points and the standard Mask R-CNN using the backbones of different capacities. As expected from a stronger method, Implicit PointRend significantly outperforms the standard Mask R-CNN when both trained with points. Moreover, we find that Implicit PointRend trained with only 10 labeled points per objects achieves the performance of the fully-supervised Mask R-CNN. Given that Mask R-CNN is still actively used in many real-world applications, this result suggests broad applicability of the new annotation form in scenarios with a constrained annotation budget.

In addition, in the appendix we report Implicit PointRend performance on COCO with mask supervision.

\begin{table}[t]
\tablestyle{9pt}{1.1}
\begin{tabular}{l|x{34} x{34}}
 method & AP ($\mathcal{M}$) & AP ($\mathcal{P}_{10}$) \\
\shline
 PointRend~\cite{kirillov2020pointrend} & 38.3 \phantom{dt{0.0}} & 35.7 \phantom{dt{0.0}} \\
 Implicit PointRend & \textbf{38.5} \dt{+0.2} & \textbf{36.9} \dt{+1.2} \\
\end{tabular}\vspace{-3mm}
\caption{\textbf{PointRend \vs Implicit PointRend} with mask ($\mathcal{M}$) and our point supervision ($\mathcal{P}_{10}$) on COCO \texttt{val2017}. A ResNet-50-FPN~\cite{he2016deep,lin2016feature} backbone is used for both models. While Implicit PointRend performs on par with the standard PointRend with the full mask supervision (AP ($\mathcal{M}$)), it significantly outperforms the baseline in case of the point supervision (AP ($\mathcal{P}_{10}$)).
}
\label{tab:coco_arch}
\end{table}

\begin{table}[t]
  \vspace{-1mm}
\centering
\tablestyle{3pt}{1.1}
\begin{tabular}{lc|x{34}x{34}x{34}}
 method & supervision &  R50 AP & R101 AP & X101 AP \\
\shline
 Mask R-CNN & $\mathcal{M}$ & \textbf{37.2} \phantom{\dt{-0.0}} & \textbf{38.6} \phantom{\dt{-0.0}} & 39.5 \phantom{\dt{-0.0}} \\
 Mask R-CNN & $\mathcal{P}_{10}$ & 36.0 \dt{-1.2} & 37.8 \dt{-0.8} & 38.5 \dt{-1.0} \\
Implicit PointRend & $\mathcal{P}_{10}$ & 36.9 \dt{-0.3} & 38.5 \dt{-0.1} & \textbf{39.7} \dt{+0.2} \\
\end{tabular}\vspace{-3mm}
\caption{Implicit PointRend performance with 10 points supervision achieves the performance of the standard Mask R-CNN model with full mask supervision on COCO \texttt{train2017}. We use R50, R101, and X101 backbones~\cite{he2016deep,xie2017aggregated} with FPN~\cite{lin2016feature}.}
\label{tab:ipr:ablation}
\end{table}

\section{Conclusion}
We present the new point annotation scheme for instance segmentation which labels only a bounding box and several random points annotated per instance. Unlike many other weak annotation forms, the resulting point-based supervision can be seamlessly applied to existing instance segmentation models without any modification to their architecture or training algorithm. The new annotation scheme provides the best trade-off between annotation time and accuracy among other schemes for instance segmentation. We further propose a simple but effective module named Implicit PointRend which tackles the unique challenges of point supervision with implicit mask representation.

\newpage
\paragraph{Acknowledgments.} We would like to thank Ross Girshick, Yuxin Wu, Piotr Doll{\'a}r, Alex Berg, Tamara Berg, and Elisa Berger for useful discussions and advices.

\appendix
\begin{center}{\bf \Large Appendix}\end{center}\vspace{-2mm}
We first provide more details for the point classification annotation tool in section~\ref{app:pipeline}. In section~\ref{app:aug}, we provide additional ablation experiments analyzing overfitting with point-based supervision as well as a more detailed study of the proposed point-based data augmentation. Section~\ref{app:coco_time} discusses annotation time for COCO dataset~\cite{lin2014coco} with different types of supervision. Finally, we conduct a thorough analysis of Implicit PointRend in section~\ref{app:ipr}.

\section{Annotation Pipeline}
\label{app:pipeline}

\begin{figure}[!t]
    \centering
    \includegraphics[width=0.99\linewidth]{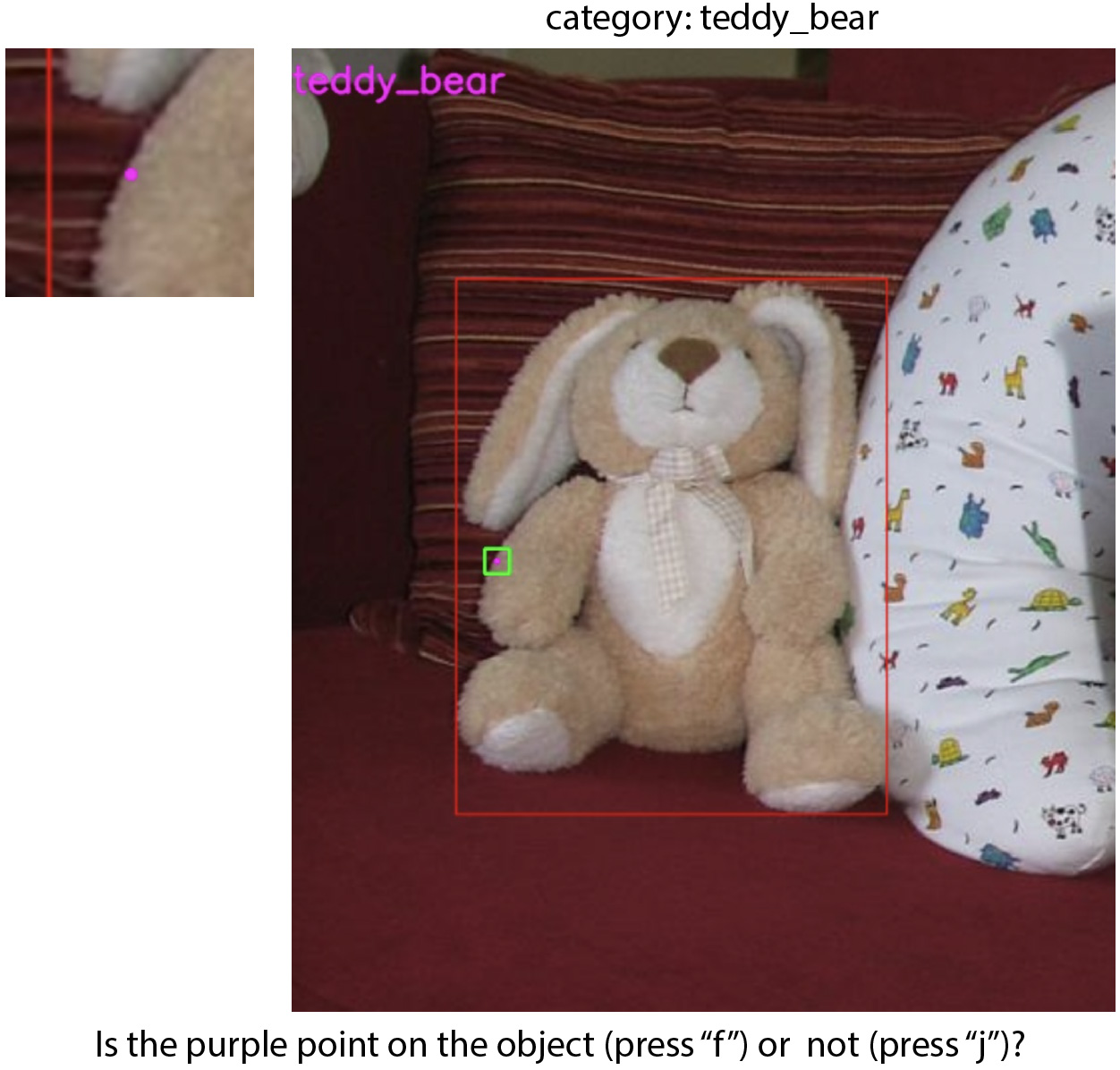}
    \caption{\textbf{Screenshot of the point annotation tool.} For each object, points are presented one by one. The tool uses two views of each point to simplify the task: (right) view shows the whole object annotated with a bounding box and a category. The point is depicted as a purple circle surrounded with a green box to simplify its spotting; For small objects, this view is cropped around the bounding box with margins for context; (left) view shows zoomed in patch of the image centered around the point to classify. This view helps to correctly classify points that are close to boundaries.
    }
\label{fig:annotation_tool}
\end{figure}

To estimate the speed and quality of the point-based annotation scheme we developed a simple labeling tool. The screenshot of its interface is shown in Figure~\ref{fig:annotation_tool}. An annotator is presented with randomly sampled points one by one for each object. Our tool shows two views for a point. The first view contains the whole object together with its bounding box, category, and a point marker. The view is centered around the object bounding box with additional margins for context. Apart from the location of the point marker, this view does not change between different points. Note, that the point marker can be small and hard to spot for large objects. To make it more visible we add a green box around the point in this view. The second view shows zoomed in area centered around the point to help classify harder cases. Such two views system allows an annotator to classify points without the need to zoom in on them manually. Our experiments with COCO data show that a trained annotator labels a point in less than a second (0.8 -- 0.9 seconds on average).

\begin{figure}[!t]
    \centering
    \includegraphics[width=0.99\linewidth]{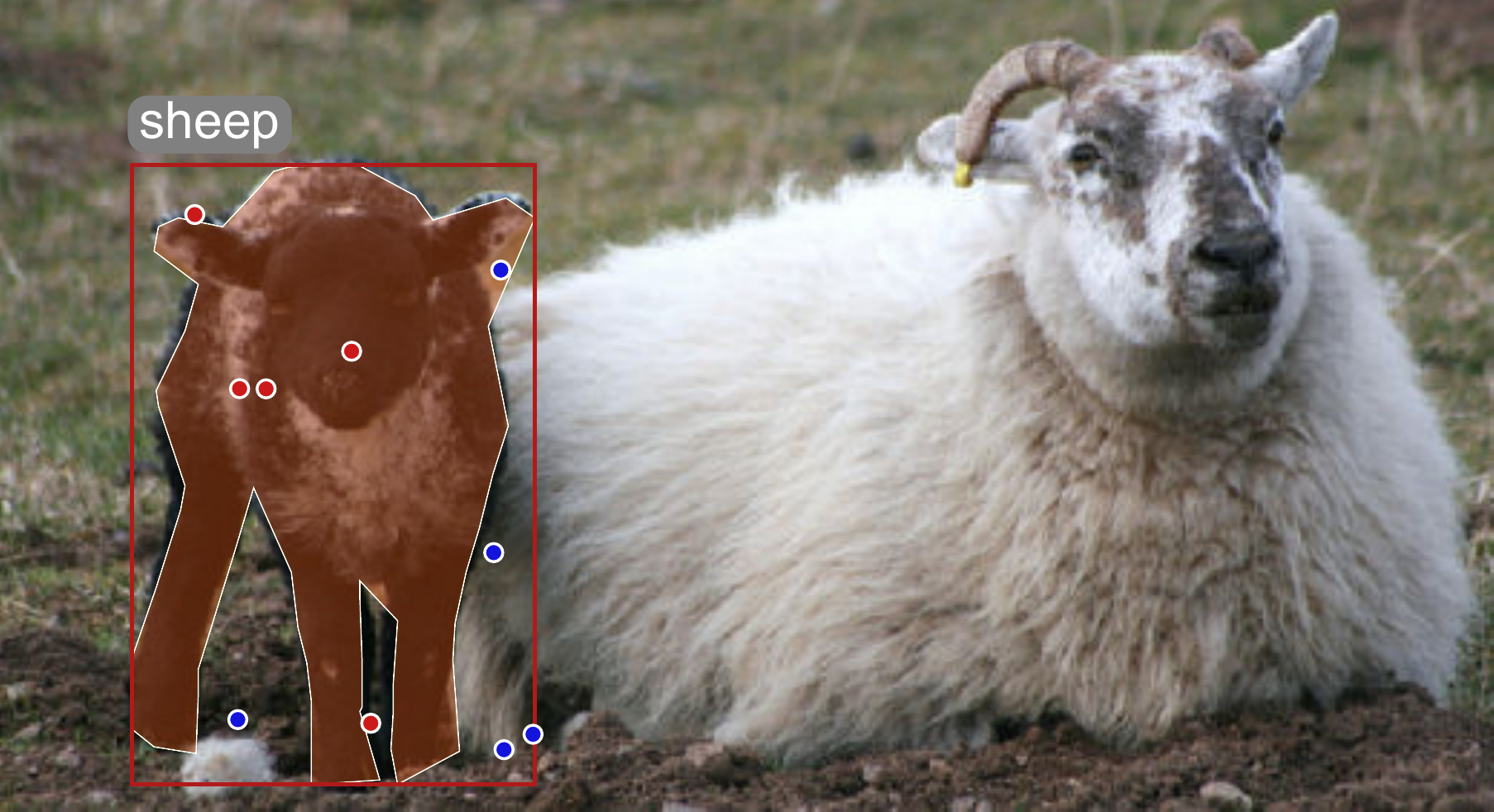}
    \caption{\textbf{Point annotation \vs polygon-based mask on COCO.} \textcolor{red}{Red} points were annotated as object and \textcolor{blue}{blue} as background. Note, that 3 points (two near both ears and one between the front legs) have correct labels but do not match to the polygon-based mask annotation. We observe that on COCO most of disagreement between point labels and ground truth masks have similar nature.
    }
\label{fig:annotation_error}
\end{figure}

We estimate the quality of our point annotation by checking its labels against ground truth masks. We observed $\sim$90\% agreement between points and instance masks on COCO. Upon closer analysis, we find that most of the errors are due to inaccurate boundaries in COCO polygon-based annotation (see Figure~\ref{fig:annotation_error}).

\begin{figure}[!t]
  \centering
  \includegraphics[width=0.89\linewidth]{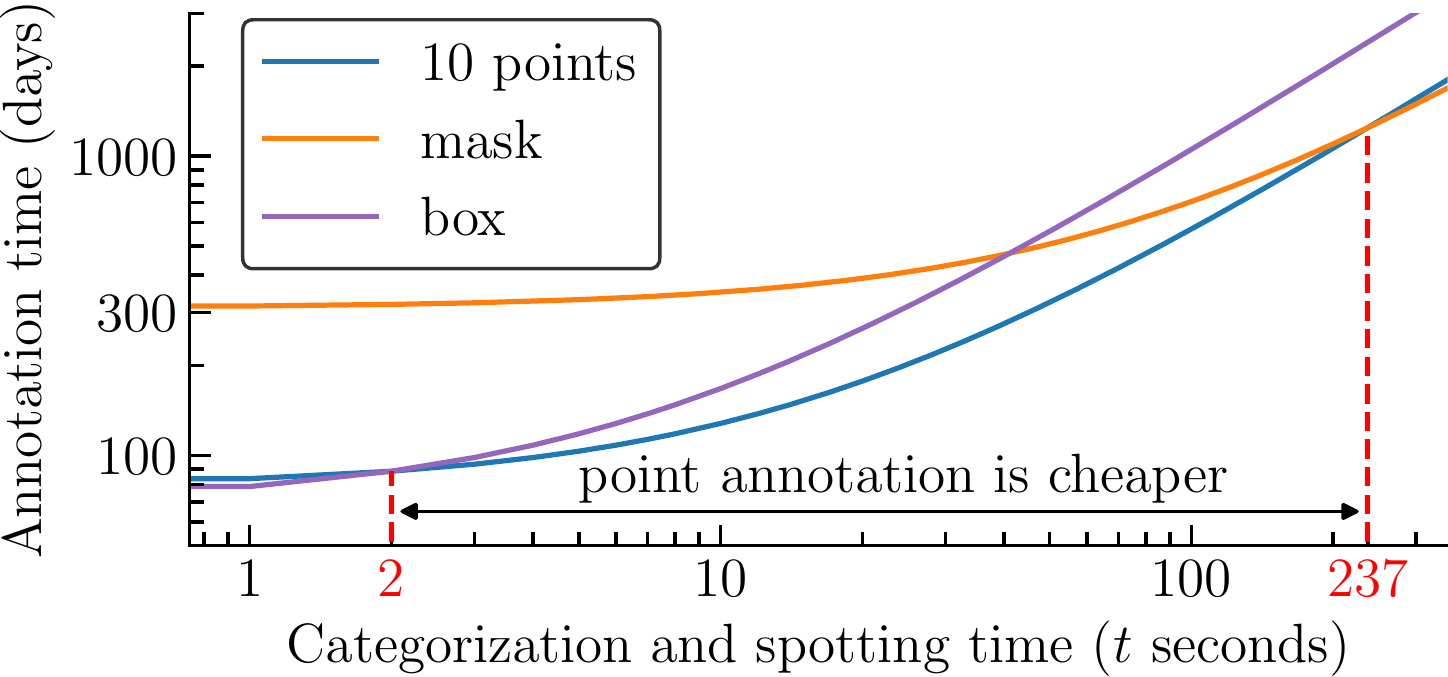}
  \vspace{-2mm}
\caption{\textbf{Total annotation time to achieve 31.8 AP on COCO} depending on the speed of categorization and spotting stages. BoxInst~\cite{tian2020boxinst} trained with boxes achieves this performance on full COCO \texttt{train2017}. Whereas, Mask R-CNN needs only $\sim$40\% and $\sim$50\% of the train set with mask and points supervision respectively. Using this data, we estimate that point-based annotation is more efficient for labeling pipelines where the spotting and categorization stages take between 2 and 237 seconds per instance.
}
\vspace{-2mm}
\label{fig:total_time_vs_t}
\end{figure}

\paragraph{Estimation for various annotation pipelines.} Different annotation collection pipelines for instance segmentation use different protocols for object spotting and categorization stages~\cite{lin2014coco,gupta2019lvis,kuznetsova2020openimages}. Note, that time $t$ spent on these stages is the same no matter what annotation form for instances is used. For the original COCO annotation pipeline, $t$ is 43.2 seconds per instance, thanks to multiple additional verification steps~\cite{lin2014coco}. In Figure~\ref{fig:total_time_vs_t} we compute total annotation time needed to achieve 31.8 AP on COCO\footnote{BoxInst~\cite{tian2020boxinst} performance with full COCO \texttt{train2017} set.} with different types of supervision for any $t \ge 0$. We observe, that the point-based annotation is more efficient for labeling pipelines where the spotting and categorization stages take between 2 and 237 seconds per instance on average.

\section{Overfitting and Point-based Augmentation}
\label{app:aug}

\paragraph{Overfitting with longer training schedules.} In our experiments with the COCO dataset~\cite{lin2016feature}, we observe that the gap between Mask R-CNN models~\cite{he2017mask} trained with full mask and point-based (10 points) supervision increases for longer training schedules. For example, for a ResNet-50-FPN~\cite{he2016deep,lin2016feature} backbone, the gap is 0.9 AP (35.2 AP \vs 34.3 AP) for 1$\x$ schedule, but grows to 1.1 AP (37.2 AP \vs 36.1 AP) for longer 3$\x$ schedule. We hypothesize that this is caused by a reduced variability of training data when only a few points are available and propose a simple \emph{point-based} data augmentation strategy to counter the effect. Note, that in our paper we train all COCO models with 3$\x$ schedule.

\paragraph{Point-based data augmentation analysis.} The proposed point-based augmentation randomly samples half of the points for a box at each iteration instead of using all of them. We vary the number of sampled points (with 1, 3, 5, 7, or 9 points) for 10 points ground truth and find that taking only 1 or 3 points is too aggressive while the results are similar when sampling 5, 7, or 9 points.

\section{More Analysis on Point-based Annotation}
\label{app:analysis}

\paragraph{Point sampling schemes.}
The advantage of using \emph{one} random point over \emph{one} click has been shown in~\cite{bearman2016s}. We are not able to make direct comparison with multiple clicks since the reliable simulation of clicks is impossible and collecting such annotation for large-scale datasets used in our draft is prohibitively expensive for an ablation. We observe that random points are already close to full supervision while being more efficient to collect and simulate. We further explore non-uniform sampling scheme by sampling more points close to boundaries on COCO. We observe the performance of a Mask R-CNN model decreases more with a heavier bias -- uniform: 36.1 AP, mildly biased to boundaries: 35.5 AP, heavily biased to boundaries: 27.5 AP.

\paragraph{Point-based annotation for rare objects.}
We re-generate $\mathcal{P}_{10}$ 3 times for LVIS dataset~\cite{gupta2019lvis} which has a lot of rare categories (occur in $<10$ images). For each version, we train a Mask R-CNN 3 times and the mean and std of AP for rare classes are: 10.8$\pm$0.85, 10.2$\pm$0.30 and 10.7$\pm$0.40. The std over 3 dataset versions is 0.26 which is smaller than the std (0.78) for model trained with mask supervision\footnote{\url{https://www.lvisdataset.org/bestpractices}}. This suggests that sampling 10 random points is sufficiently robust for rare categories.

\section{COCO Annotation Time}
\label{app:coco_time}

The annotation protocol of the COCO dataset~\cite{lin2014coco} has 3 stages: (1) category labeling, (2) instance spotting and (3) mask annotation. The creators of the COCO dataset report detailed timing of each stage on all annotated images. Here, we summarize the annotation time \emph{per instance} for different types of supervision.

\paragraph{Category labeling and instance spotting.} Category labeling is the task of determining which object categories are present in each image. This step takes around 7,000 hours to annotate 118,287 images with 849,949 instances (\texttt{train2017} set), suggesting the time of category labeling is $28.8$ seconds per instance. The instance spotting stage places a cross on top of each instance. This step takes 3,500 hours to annotate 118,287 images with 849,949 instances. Thus, instance spotting takes $14.4$ seconds per instance. Note that, in order to achieve a high recall, both category labeling and instance spotting are performed with 8 workers for every image and the total annotation time mentioned before is the sum of 8 workers.

\paragraph{Bounding box annotation.} Since the original COCO protocol does not annotate instances by bounding boxes, we approximate the time of bounding box annotation ($\mathcal{B}$) as 7 seconds for a spotted instance which can be achieved with extreme clicks technique~\cite{papadopoulos2017extreme}. Taken into account category labeling and instance spotting, bounding box annotation ($\mathcal{B}$) takes $28.8 \text{ (category labeling)} + 14.4 \text{ (instance spotting)} + 7.0 \text{ (box)} = 50.2$ seconds per instance.

\paragraph{Mask annotation.} Polygon-based mask annoation in COCO takes 22 hours per 1,000 instances or $\sim$79.2 seconds per instance. Together with category labeling and instance spotting, mask annotation ($\mathcal{M}$) takes $28.8 \text{ (category labeling)} + 14.4 \text{ (instance spotting)} + 79.2 \text{ (mask)} = 122.4$ seconds per instance.

\paragraph{Our point-based annotation.} Given the bounding box, it takes annotator 0.8 -- 0.9 seconds on average to provide the binary label for each point. Thus, our point-based annotation with 10 points ($\mathcal{P}_{10}$) takes $50.2 \text{ (bounding box annotation)} + 0.9 \times 10 \text{ (10 points per instance)} = 59.2$ seconds per instance which is more than 2 times faster than mask annotation if the time of categorization and spotting stages is included. For datasets with bounding box annotations~\cite{shao2019objects365,kuznetsova2020openimages}, our point annotations takes only $0.9 \times 10 (\text{10 points per instance}) = 9$ seconds per instance which is more than 8 times faster than the polygon-based mask annotation ($79.2$ seconds per instance). In Table~\ref{tab:anno_time_coco}, we report the annotation time for different annoation format on COCO \texttt{train2017} set, which contains 118,287 images and 849,949 instances.

\begin{table}[t]
    \tablestyle{4pt}{1.1}
    \begin{tabular}{lc|x{110}}
     supervision & notation & COCO annotation time (days)\\
    \shline
     bounding box & $\mathcal{B}$ & \phantom{0}493 \\
     mask & $\mathcal{M}$ & 1204 \\
     our point-based & $\mathcal{P}_{10}$ & \phantom{0}582 \\
    \end{tabular}\vspace{2mm}
    \caption{\textbf{Annotation times for different types of supervision} on COCO \texttt{train2017}. We report the time to annotate 118,287 images with 849,949 instances. $\mathcal{B}$: bounding box supervision, $\mathcal{M}$: mask supervision, and $\mathcal{P}_{10}$: our point-based supervision with a bounding box and 10 points labels per instance.}
    \label{tab:anno_time_coco}
\end{table}

\begin{table}[t]
    \centering
    \tablestyle{2pt}{1.1}
    \begin{tabular}{x{32}|x{60}x{55}x{70}}
     point aug. & Mask R-CNN~\cite{he2017mask} & PointRend~\cite{kirillov2020pointrend} & Implicit PointRend \\
    \shline
     \hline
     & 36.1 \phantom{\dt{-0.0}} & 35.6 \phantom{\dt{+0.0}} & 36.0 \phantom{\dt{+0.0}} \\
     \checkmark & 36.0 \dt{-0.1} & 35.7 \dt{+0.1} & 36.9 \dt{+0.9} \\
    \end{tabular}\vspace{2mm}\\
    \caption{\textbf{Point-based augmentation for various models} with a ResNet-50-FPN~\cite{he2016deep,lin2016feature} backbone. All model are trained with 10 points supervision on COCO \texttt{train2017} and mask AP on \texttt{val2017} is reported. The new augmentation is more effective for Implicit PointRend as the new module has higher capacity in representing object masks with its parameter head.}
    \label{tab:ablation_sup:augmentation}
\end{table}

\section{Implicit PointRend}
\label{app:ipr}

In this section, we ablate the design of the Implicit PointRend module and report its performance with full mask supervision for reference.

\subsection{Ablation study}
\label{app:ipr:ablation}

For the ablation experiments we use a ResNet-50~\cite{he2016deep} backbone with FPN~\cite{lin2016feature}, $3\times$ schedule and point-based data augmentation.

\paragraph{Point-based data augmentation.} We observe that point-based data augmentation does not significantly improve performance for Mask R-CNN~\cite{he2017mask} and PointRend~\cite{kirillov2020pointrend} with a ResNet-50-FPN backbone (see Table~\ref{tab:ablation_sup:augmentation}). Whereas, Implicit PointRend shows 0.9 AP gain with the augmentation. We hypothesize that the augmentation is more effective for Implicit PointRend as the new module has a higher capacity in representing object masks with its parameter head.

\begin{table}[t]
    \centering
    \subfloat[\textbf{Coordinates.}\label{tab:ablation_sup:coord}]{
    \tablestyle{2pt}{1.1}
    \begin{tabular}{l|x{34}x{34}}
     coord. &  AP ($\mathcal{P}_{10}$) & AP ($\mathcal{M}$) \\
    \shline
     none & 35.2 \phantom{\dt{+0.0}} & 37.7 \phantom{\dt{+0.0}} \\
     rel. & 36.6 \dt{+1.4} & 38.5 \dt{+0.8} \\
     p.e. & 36.9 \dt{+1.7} & 38.5 \dt{+0.8} \\
    \end{tabular}}\hspace{5mm}
    \subfloat[\textbf{Image-level features.}\label{tab:ablation_sup:feature}]{
    \tablestyle{2pt}{1.1}
    \begin{tabular}{l|x{34}x{34}}
     features &  AP ($\mathcal{P}_{10}$) & AP ($\mathcal{M}$) \\
    \shline
     none & 35.3 \phantom{\dt{+0.0}} & 37.1 \phantom{\dt{+0.0}} \\
     \texttt{p2} & 36.9 \dt{+1.6} & 38.5 \dt{+1.4} \\
     \multicolumn{2}{c}{~} \\
    \end{tabular}}\vspace{2mm}\\
    
    \caption{\textbf{Implicit PointRend ablations} on COCO \texttt{train2017} with ResNet-50-FPN~\cite{he2016deep,lin2016feature} backbone. AP ($\mathcal{P}_{10}$) is mask AP for point-based supervision and AP ($\mathcal{M}$) is mask AP for full mask supervision. Table~\ref{tab:ablation_sup:coord}: Implicit PointRend utilizes the relative coordinate \wrt the box point-level information~(\emph{rel.}) to improve instance segmentation performance. Positional encoding representation for the coordinate~(\emph{p.e.})~\cite{tancik2020fourier} further improve results. Table~\ref{tab:ablation_sup:feature}: Implicit PointRend can achieve reasonable performance with only the positional encoding as the input to its point head~(\emph{none}). Image-level point features from the \texttt{p2} level of FPN~\cite{lin2016feature} backbone~(\emph{p2}) further improve the performance
    }
    \label{tab:ablation_sup}
    \end{table}

\paragraph{Point-level feature representation.} The point head of Implicit PointRend takes two types of features as input: point coordinate relative to the bounding box and image-level point features. Next, we ablate both components.

In Table~\ref{tab:ablation_sup:coord}, we show that adding the relative coordinate \wrt the box is already effective, giving an improvement of $1.4$ AP with point-based supervision and $0.8$ AP with mask supervision. Encoding the coordinate with a positional encoding~\cite{tancik2020fourier} further improves the mask prediction by $0.3$ AP for point-based supervision. Without the coordinates, the mask head of Implicit PointRend is translation invariant and cannot distinguish two points with similar appearance in the same bounding box. PointRend~\cite{kirillov2020pointrend} breaks the invariance by taking coarse mask prediction as input.

Table~\ref{tab:ablation_sup:feature} shows that Implicit PointRend can achieve reasonable performance with only the positional encoding as the input to its point head. Image-level point features from the \texttt{p2} level of FPN~\cite{lin2016feature} backbone further improve the performance by $1.6$ AP with point-based supervision and $1.4$ AP with mask supervision. These experiments suggest that both coordinate and image-level features are essential for the overall performance of Implicit PointRend.

\subsection{Mask supervision results}
\label{app:ipr:sota}
We compare Implicit PointRend with several instance segmentation approaches on the COCO dataset~\cite{lin2014coco} with full mask supervision. The module design is motivated by our point-based supervision, however, Table~\ref{tab:coco_full} shows that Implicit PointRend is also a competitive method for fully-supervised instance segmentation as well. While both CondInst~\cite{tian2020conditional} and Implicit PointRend use an instance-dependent function to predict masks, Implicit PointRend outperforms CondInst by a large margin. Implicit PointRend is able to match the performance of PointRend~\cite{kirillov2020pointrend} without the coarse mask prediction and importance sampling procedure during training. We note that the small gap (less than $0.3$ AP) between Implicit PointRend and PointRend mostly comes from large objects. We hypothesize that the fixed-size feature map (\eg, $14\times14$) extracted from \texttt{p2} FPN level is too coarse to generate point head parameters to accurately segment large objects. A simple fix could be dynamically choosing the FPN levels to pool feature like the box head~\cite{lin2016feature}. In our experiments such design gave a small 0.1-0.2 AP boost for Implicit PointRend results. We expect a better design for the parameter head can further improve Implicit PointRend performance. Moreover, PointRend is trained with a more complex importance sampling while much simpler uniform sampling is used for Implicit PointRend, which may also lead to the small drop when larger backbones are used.

\begin{table}[t!]
    \centering
    \vspace{3.5mm}
    \tablestyle{1.0pt}{1.1}
    \begin{tabular}{l c c|x{22}x{22}x{22}|x{22}x{22}x{22}}
     & backb. & LRS & AP &  AP$_{50}$ & AP$_{75}$ & AP$_S$ &  AP$_M$ &  AP$_L$\\
    \shline \rule{0mm}{5mm}
     \multirow{4}{49pt}{Mask R-CNN~\cite{he2017mask}} & R50 & $1\x$ & 35.2 & 56.3 & 37.5 & 17.2 & 37.7 & 50.3 \\
      & R50 & $3\x$ & 37.2 & 58.6 & 39.9 & 18.6 & 39.5 & 53.3 \\
      & R101 & $3\x$ & 38.6 & 60.4 & 41.3 & 19.5 & 41.3 & 55.3 \\
      & X101 & $3\x$ & 39.5 & 61.7 & 42.6 & 20.7 & 42.0 & 56.5 \\[2.5mm]
    \hline \rule{0mm}{5mm}
     \multirow{3}{49pt}{CondInst~\cite{tian2020conditional}} & R50 & $1\x$ & 35.7 & - & - & - & - & - \\
      & R50 & $3\x$ & 37.5 & - & - & - & - & - \\
      & R101 & $3\x$ & 38.6 & - & - & - & - & - \\[2.5mm]
    \hline \rule{0mm}{5mm}
     \multirow{4}{49pt}{PointRend~\cite{kirillov2020pointrend}} & R50 & $1\x$ & 36.2 & 56.6 & 38.6 & 17.1 & 38.8 & 52.5 \\
      & R50 & $3\x$ & 38.3 & 59.1 & 41.1 & 19.1 & 40.7 & 55.8 \\
      & R101 & $3\x$ & 40.1 & 61.1 & 43.0 & 20.0 & 42.9 & 58.6 \\
      & X101 & $3\x$ & 41.1 & 62.8 & 44.2 & 21.5 & 43.8 & 59.1 \\[2.5mm]
    \hline \rule{0mm}{5mm}
     \multirow{4}{49pt}{Implicit PointRend} & R50 & $1\x$ & 36.9 & 57.3 & 39.6 & 17.7 & 39.4 & 53.1 \\
      & R50 & $3\x$ & 38.5 & 59.4 & 41.4 & 18.9 & 41.1 & 55.0 \\
      & R101 & $3\x$ & 39.9 & 61.1 & 43.0 & 20.3 & 42.7 & 58.0 \\
      & X101 & $3\x$ & 40.8 & 62.6 & 43.9 & 21.5 & 43.5 & 57.3 \\
    \end{tabular}\vspace{2mm}
    \caption{\textbf{Instance segmentation results with full mask supervision} on COCO \texttt{val2017}. LRS: learning rate schedule; a $1\x$ learning rate schedule refers to 90,000 iterations. R50,101: ResNet-50,101~\cite{he2016deep}. X101: ResNext-101 $32\x8$d~\cite{xie2017aggregated}. All models use FPN~\cite{lin2016feature}. The proposed Implicit PointRend also performs better than Mask R-CNN~\cite{he2017mask} and comparable to PointRend~\cite{kirillov2020pointrend} under full mask supervision.}
    \label{tab:coco_full}
    \end{table}

{\small
\bibliographystyle{ieee_fullname}
\bibliography{egbib}
}

\end{document}